\documentclass[10pt,twocolumn,letterpaper]{article}

 \usepackage[pagenumbers]{cvpr} 

\definecolor{cvprblue}{rgb}{0.21,0.49,0.74}
\usepackage[pagebackref,breaklinks,colorlinks,allcolors=cvprblue]{hyperref}
\usepackage{algorithm}
\usepackage{algorithmic}

\usepackage{amsfonts}
\usepackage{booktabs}
\usepackage{multirow}
\usepackage{pifont}
\usepackage{color}
\usepackage{colortbl}
\usepackage{tabularx}
\usepackage{array} 
\usepackage{xcolor}
\usepackage{amsmath}

\definecolor{best}{rgb}{1, 0.5, 0}
\definecolor{second}{rgb}{1, 1, 0}

\def\paperID{} 
\def\confName{CVPR Workshop}
\def\confYear{2025}

\title{SMURF: Continuous Dynamics for Motion-Deblurring Radiance Fields\vspace{-4mm}}

\author{{\fontsize{11}{11}\selectfont Jungho Lee \quad Dogyoon Lee \quad Minhyeok Lee \quad Donghyeong Kim \quad Sangyoun Lee}\vspace{2mm}\\
	{\fontsize{11}{11}\selectfont School of Electrical and Electronic Engineering, Yonsei University} \\
	{\fontsize{10}{10}\selectfont \tt \textbf{\href{https://Jho-Yonsei.github.io/SMURF/}{\texttt{https://Jho-Yonsei.github.io/SMURF/}}}}
}

\begin{document}
\maketitle
\begin{abstract}
Neural radiance fields (NeRF) has attracted considerable attention for their exceptional ability in synthesizing novel views with high fidelity. However, the presence of motion blur, resulting from slight camera movements during extended shutter exposures, poses a significant challenge, potentially compromising the quality of the reconstructed 3D scenes. To effectively handle this issue, we propose sequential motion understanding radiance fields (SMURF), a novel approach that models continuous camera motion and leverages the explicit volumetric representation method for robustness to motion-blurred input images. The core idea of the SMURF is continuous motion blurring kernel (CMBK), a module designed to model a continuous camera movements for processing blurry inputs. Our model is evaluated against benchmark datasets and demonstrates state-of-the-art performance both quantitatively and qualitatively.
\end{abstract}  
\section{Introduction}
\label{sec:intro}


Generally, NeRF variants have reconstructed 3D scenes using well-captured, noise-free images as inputs along with calibrated camera parameters. The training of complex geometry in 3D scenes necessitates sharp images; however, in real-world scenarios, obtaining such images may not always be feasible due to various factors. For instance, to capture a sharp image, it is essential to set the camera's aperture to a small size, thereby increasing the depth of field. However, a smaller aperture demands a significant amount of light, which consequently leads to longer exposure times. During these extended exposure, any movement of the handheld camera results in undesirable camera motion-blurred images. Recently, many works~\cite{ma2022deblurnerf,lee2023dp,wang2023bad,peng2023pdrf,lee2023exblurf} have been conducted on NeRF that takes camera motion-blurred images as input. Deblur-NeRF~\cite{ma2022deblurnerf} first proposes a method that models blurring kernel by imitating the deconvolution method for blind image deblurring, to reconstruct 3D scenes from motion-blurred images and render sharp novel view images. Since the inception of Deblur-NeRF, various methods~\cite{lee2023dp,wang2023bad,peng2023pdrf,lee2023exblurf} for estimating the blurring kernel have been proposed. However, their blurring kernel does not continuously model the camera movement during exposure time. Since actual camera movement is normally continuous, it can be represented as a continuous function over time, and such continuous modeling allows for a more precise tracking of the camera movement path, even when the movement is complex or irregular. Previous works model the blurring kernel without considering the sequential camera movement, making the kernel for images with complex camera motion imprecise.

\begin{figure}[t]
	\centering
	\includegraphics[width=\linewidth]{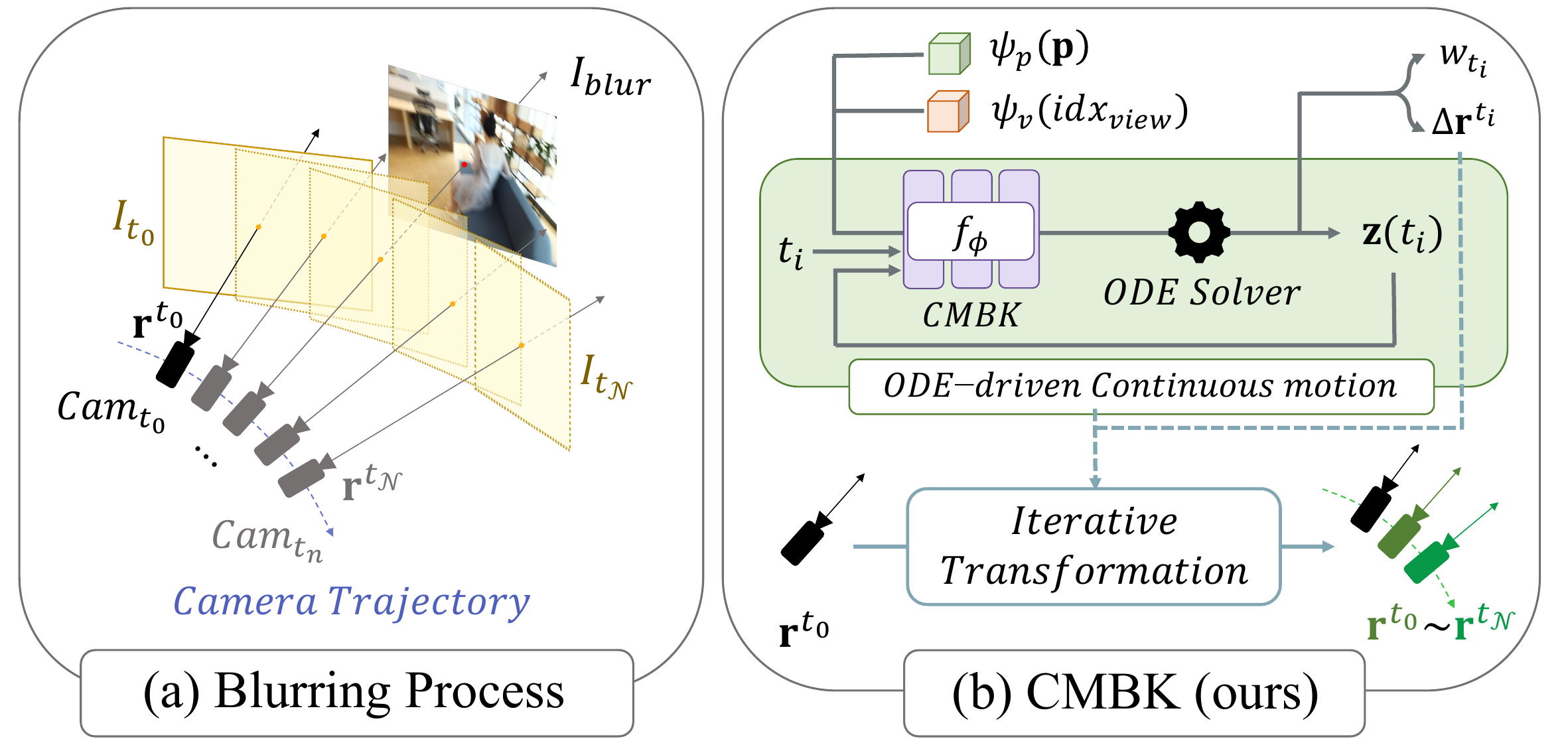}
	\caption{\textbf{Blurring process and our sequential kernel generation.} A blurry image $I_{blur}$ is acquired as the camera moves over the exposure time, with images $I_{t}$ captured at each camera pose being composited together. Our approach estimates warped rays along the continuous camera motion trajectory.}
	\label{fig:deblur_vs_smurf}
	\vspace{-4mm}
\end{figure}

In this paper, we propose a sequential motion understanding radiance fields (SMURF), which incorporates a novel continuous motion blur kernel (CMBK) for modeling precise camera movements from blurry images. The CMBK estimates the small change in pose regarding continuous camera motion. The output values of the CMBK are not computed jointly as in previous studies but are designed to be computed sequentially according to camera motion, as shown in \cref{fig:deblur_vs_smurf}. 
Additionally, we adopt explicit volumetric representation methods~\cite{jun2022hdrplenoxel,chen2022tensorf,muller2022instant,sun2022direct} as its backbone. Specifically, the tensor factorization-based approach, Tensorial Radiance Fields (TensoRF)~\cite{chen2022tensorf}, facilitates compact and efficient 3D scene reconstruction. 

To demonstrate the effectiveness of the proposed SMURF, we conduct extensive experiments on synthetic and real-world scenes. Our experimental results elucidate the advantages of the CMBK in comparison to previously presented blurring kernels. Our main contributions are summarized as follows:
\begin{itemize}
	\item [$\bullet$] We propose a continuous motion blur kernel (CMBK) to sequentially estimate continuous camera motion from blurry images mimicking the real-world motion blur.
	\item [$\bullet$] Our sequential motion understanding radiance fields (SMURF) exhibit higher visual quality and quantitative performance compared to existing approaches.
\end{itemize}
\section{Preliminary}
\label{sec:pre}

\paragraph{\textbf{Blind Deblurring for 3D Scene. }}
\label{subsubsec:deblurring}

Deblur-NeRF~\cite{ma2022deblurnerf} apply the image blind deblurring algorithm to NeRF, modeling an adaptive blurring kernel for each ray to get the blurry pixel. Deblur-NeRF designs the sprase kernel to acquire blurry color, and for the temporally continuous camera motion, we extend the process as follows:
\begin{equation} \label{eq:nerf_blur}
	\mathbf{c}_{blur}(\mathbf{r}) = \sum_{i=0}^{\mathcal{N}} w_{p}^{t_{i}}\mathbf{c}_{p}(\mathbf{r}^{{t}_{i}}),~w.r.t.~\sum_{i=0}^{\mathcal{N}}w_{p}^{t_{i}}=1,
\end{equation}
where $\mathcal{N}$ and $t_{i}$ respectively denote the number of warped rays in camera motion and the instantaneous time during exposure; $w_{p}$ is the corresponding weight at each ray's location, and $\mathbf{r}^{t_{i}}$ represents the warped ray.

\paragraph{\textbf{Tensorial Radiance Fields (TensoRF)}. }
\label{subsubsec:tensorf}
We follow the architecture of TensoRF~\cite{chen2022tensorf}, an explicit voxel-based volumetric representation utilizing CP~\cite{carroll1970analysis,harshman1970foundations} and block term decomposition~\cite{de2008decompositions}, which models an efficient view-dependent sparse voxel grid. TensoRF optimizes two 3D grid tensors, $\mathcal{G}_{\sigma}, \mathcal{G}_\mathbf{c} \in \mathbb{R}^{FXYZ}$, for estimating volume density and view-dependent appearance feature, employing the vector-matrix (VM) decomposition that effectively extends CP decomposition. 
The voxel grid is represented as follows:
\begin{equation}
	\mathcal{G}=\sum_{r=1}^{R_1} \mathbf{v}_r^X \circ \mathbf{M}_r^{YZ}+\sum_{r=1}^{R_2} \mathbf{v}_r^Y \circ \mathbf{M}_r^{XZ}+\sum_{r=1}^{R_3} \mathbf{v}_r^Z \circ \mathbf{M}_r^{XY},
\end{equation}
where $\circ$ denotes the outer product; $R_1, R_2$, and $R_3$ indicates the number of low-rank components. $\mathbf{v}^X \in \mathbb{R}^{X}$ is the vector along the $X$ axes in each mode, and $\mathbf{M}^{YZ} \in \mathbb{R}^{YZ}$ is the matrix in the $YZ$ plane for each mode. Once the grids are defined, the radiance field at a 3D point $\mathbf{x}$ for computing volume density~$\sigma$ and color~$\mathbf{c}$ is defined as follows:
\begin{equation}
	\sigma(\mathbf{x}), \mathbf{c}(\mathbf{x}) = \mathcal{G}_{\sigma}(\mathbf{x}), \mathcal{S}(\mathcal{G}_\mathbf{c}(\mathbf{x}), d),
\end{equation}
where $\mathcal{S}$ denotes a parameterized shallow MLP that converts viewing direction $d$ and appearance feature $\mathcal{G}_\mathbf{c}(\mathbf{x})$ into color. The features obtained from the grid, $\mathcal{G}_{\sigma}(\mathbf{x})$ and $\mathcal{G}_{\mathbf{c}}(\mathbf{x})$, are trilinearly interpolated from adjacent voxels.

To render an image for a given view, TensoRF uses a differentiable classic volume rendering technique~\cite{kajiya1984raytracing} with $\sigma$ and $\mathbf{c}$. The color $\mathbf{c}_{p}$ for each pixel reached by ray $\mathbf{r}$ is computed as follows:
\begin{equation}
	\mathbf{c}_{p}(\mathbf{r})=\sum_{i=1}^{N} T_{i}\left(1-\exp \left(-\sigma_i \delta_i\right)\right) \mathbf{c}_i,
\end{equation}
where $T_i=\exp \left(-\sum_{j=1}^{i-1} \sigma_j \delta_j\right)$ is the transmittance; $N$ and $\delta$ denote the number of sampled points and the step size between adjacent samples on ray $\mathbf{r}$, respectively.


\begin{figure*}[t]
	\centering
	\includegraphics[width=0.9\linewidth]{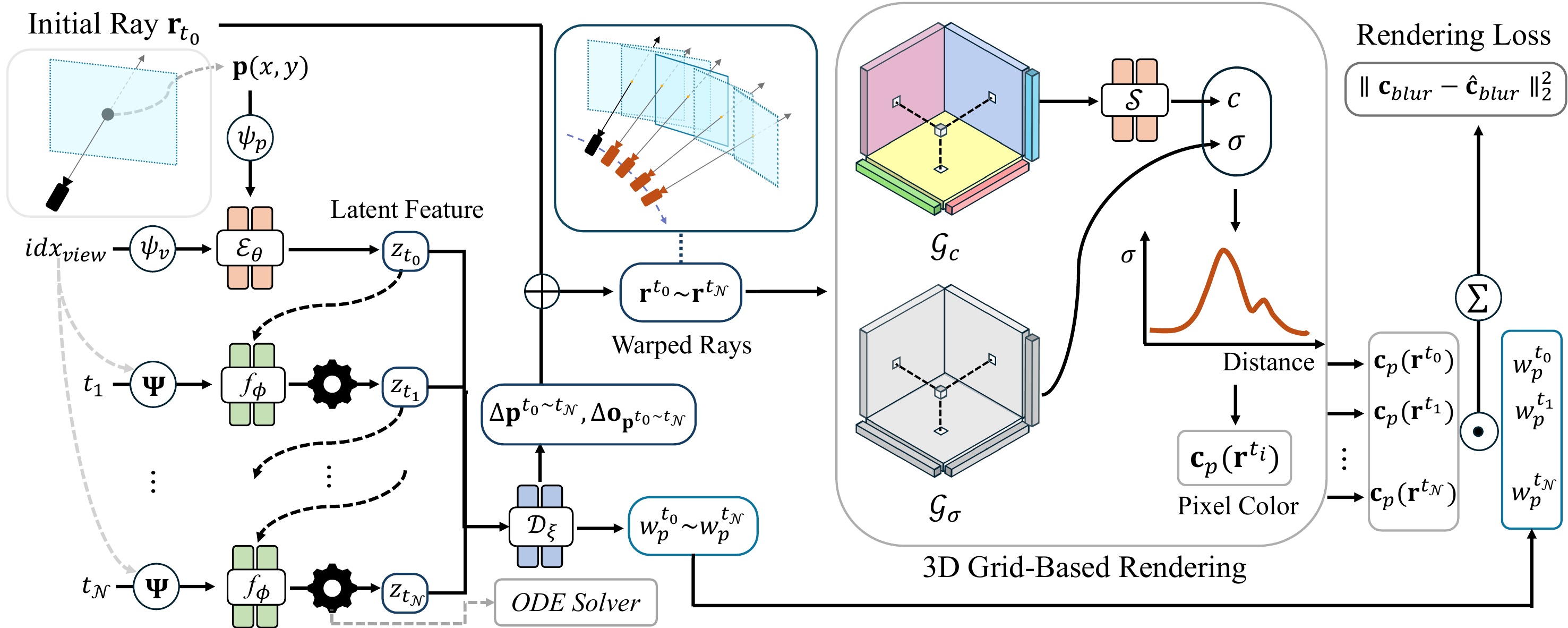}
	\caption{Method overview of the SMURF. The latent feature of initial ray is extended into an IVP with a parameterized derivative function $f_{\phi}$. Then, it is solved by Neural-ODEs along with given time~$t$, obtaining latent features for all warped rays.}
	\label{fig:method_overview}
	\vspace{-5mm}
\end{figure*}

\section{Method}
\label{sec:method}

\subsection{Continuous Motion Blur Kernel}
\label{sec:cmbk}

\paragraph{\textbf{Continuous Latent Space Modeling for Camera Motion. }}

Our goal is to model the continuous movement of the camera within the exposure time $(t_0 \leq t < t_\mathcal{N})$. To reflect the physics inherent in camera motion, we apply continuous dynamics to the latent space of camera motion, and transform the latent feature into changes in ray within the physical space. For this process, we transform the given information of the rays into latent features and apply them to a Neural-ODEs~\cite{chen2018neural} to design a continuous latent space.

As shown in \cref{fig:method_overview}, we embed information about the initial ray into the latent space using a parameterized encoder $\mathcal{E}_{\theta}$, utilizing the scene's view index $idx_{view}$ and 2D pixel location $\mathbf{p}$:
\begin{equation}
	\mathbf{z}(t_{0}) = \mathcal{E}_{\theta}\left(\psi_{v}(idx_{view}),~\psi_{p}(\mathbf{p})\right),
\end{equation}
where $\mathbf{z}(t_{0})\in\mathbb{R}^{d}$ is the latent feature of initial ray with hidden dimension $d$, and $\psi_{v}$ and $\psi_{p}$ denote embedding functions for view and pixel information, respectively. Within a small step limit of the latent feature $\mathbf{z}(t)$, the local continuous dynamic is expressed as $\mathbf{z}(t+\epsilon)=\mathbf{z}(t)+\epsilon\cdot d\mathbf{z}/dt$. To apply continuous dynamics to $\mathbf{z}(t)$, we model a parameterized neural derivative function $f$ in the latent space. The derivative of the continuous function is defined as follows: 
\begin{equation}
	\frac{d \mathbf{z}(t)}{d t}=f(\mathbf{z}(t), t; \phi),
\end{equation}
where $\phi$ denotes the learnable parameters of $f$. With $\mathbf{z}(t_{0})$ and the derivative function $f$, we define an initial value problem (IVP), and the features of subsequential rays in the latent space are obtained by the integral of $f$ from $t_{0}$ to the desired time. This dynamic leads to the format of ODEs, and the process of obtaining latent features of the camera motion trajectory using various ODE solvers~\cite{euler1845institutionum,kutta1901beitrag,dormand1980family} is expressed as follows:
\begin{equation}
	\mathbf{z}\left(t_{n+1}\right)=\mathbf{z}\left(t_n\right)+\int_{t_n}^{t_{n+1}} f(\mathbf{z}(t), t ; \phi) d t,
\end{equation}
where $0 \leq n < \mathcal{N}$, and $\mathcal{N}$ is the number of rays in the camera motion. As a result of this approach, we obtain the latent features $\mathbf{Z} \in \mathbb{R}^{\mathcal{N} \times d}$ for $\mathcal{N}$ rays.

Inspired by the difference in exposure time for images, we define a chrono-view embedding function $\mathbf{\Psi}$ with a single-layer MLP that simultaneously embeds given time $t$ and view index $idx_{view}$. Then, $f$ is expressed as follows:
\begin{equation}
	f(\mathbf{z}(t), t; \phi) \rightarrow f(\mathbf{z}(t), t, \mathbf{\Psi}(t;idx_{view}); \phi).
\end{equation}

\paragraph{\textbf{Motion-Blurring Kernel Generation. }}

The latent features $\mathbf{Z}$ of all rays on the camera motion trajectory, obtained through continuous dynamics modeling, must be decoded into changes in camera pose. We define a decoder $\mathcal{D}_{\xi}$, represented by a shallow MLP parameterized by~$\xi$, which outputs three components: the change in 2D kernel location $\Delta\mathbf{p}$, the change in ray origin $\Delta\mathbf{o}_{\mathbf{p}}$, and the corresponding weight $w_{p}$ as defined in \cref{eq:nerf_blur}:
\begin{equation}
	(\Delta\mathbf{p}^{t}, \Delta\mathbf{o}_{\mathbf{p}^{t}}, w^{t}_{p}) = \mathcal{D}_{\xi}(\mathbf{z}(t)).
\end{equation}
Then, $t$-th warped ray $\mathbf{r}^{t}$ is generated from the initial ray $\mathbf{r}=\mathbf{o}+\tau\mathbf{d}$ by following equation:
\begin{equation} \label{eq:warp}
	\mathbf{r}^{t} = (\mathbf{o} + \Delta{\mathbf{o}_{\mathbf{p}^{t}}}) + \tau\mathbf{d}_{\mathbf{p}^{t}},~\mathbf{p}^{t} = \mathbf{p} + \Delta\mathbf{p}^{t},
\end{equation}
where $\mathbf{o}$ and $\mathbf{p}$ denote the ray origin and the 2D pixel location of initial ray, respectively; $\mathbf{d}_{\mathbf{p}^{t}}$ stands for the warped direction by $\mathbf{p}^{t}$. After acquiring sharp pixel colors $\mathbf{c}_{p}(\mathbf{r}^{t})$ for all $\mathcal{N}$ warped rays with inherent continuous camera movement, the blurry pixel color is computed using~\cref{eq:nerf_blur}.

\paragraph{\textbf{Residual Momentum. }}

Latent features of the camera motion trajectory are predicted by CMBK and are expected to be optimized continuously. In this process, predicted origin and direction of the ray $\mathbf{r}^{t}$ may diverge from the initial ray $\mathbf{r}$, potentially leading to a suboptimal kernel. 
To address this issue, we apply a residual term to $f$, implementing regularization that ensures the predicted ray $\mathbf{r}^{t_{i}}$ does not significantly deviate from the previous ray $\mathbf{r}^{t_{i-1}}$:
\begin{equation}
	f_{\phi}(\mathbf{z}(t_{i-1})) \rightarrow \Lambda(f_{\phi}(\mathbf{z}(t_{i-1})) + \mathbf{z}(t_{i-1})),
\end{equation}

\begin{figure*}[t]
	\centering
	\includegraphics[width=\linewidth]{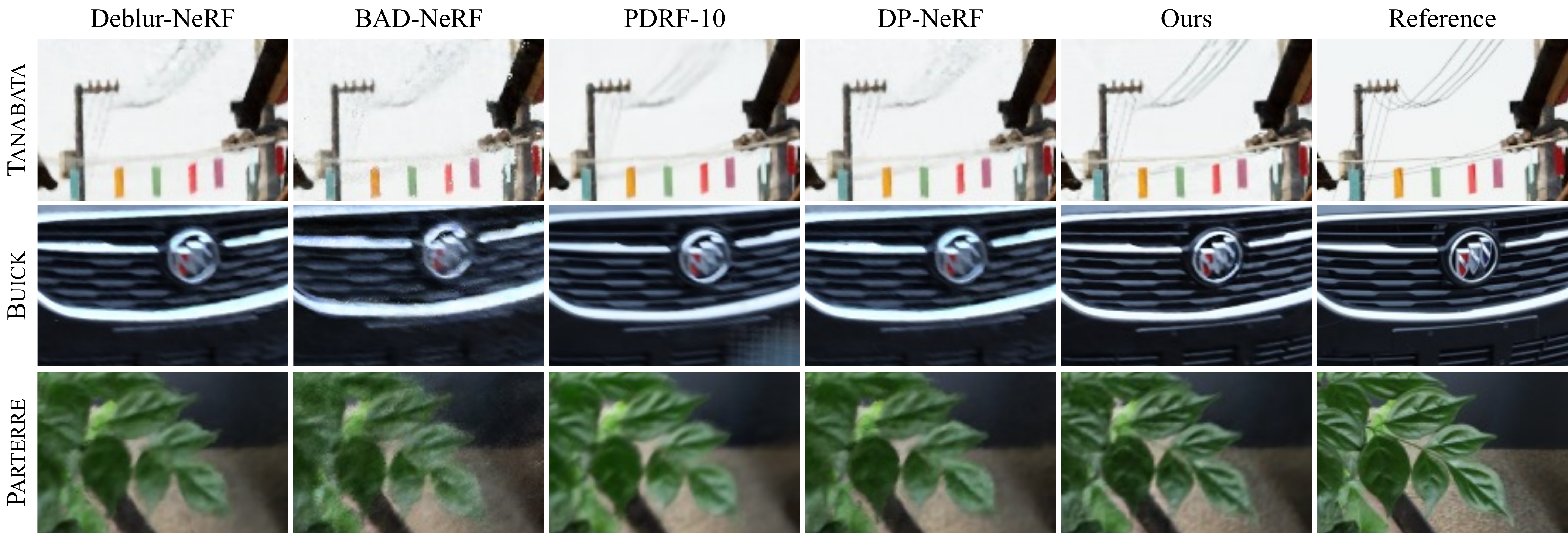}
	\vspace{-5mm}
	\caption{Qualitative comparisons on synthetic scenes and real-world scenes.}
	\label{fig:novel_view}
	\vspace{-4mm}
\end{figure*}

\paragraph{\textbf{Output Suppression Loss.}}

We encode the ray information into the latent space, and decode it into changes of the ray in motion trajectory using $\mathcal{D}_{\xi}$. In this process, since initial latent feature $\mathbf{z}(t_{0})$ serves as the initial value for the ODE, there should be no change in the initial ray. Therefore, we apply an output suppression loss as follows:
\begin{equation} \label{eq:loss_supp}
	\mathcal{L}_{supp} = \lambda_{supp}\left\|\mathcal{D}_{\xi}(\mathbf{z}(t_{0}))\right\|_{2},
\end{equation}
where $\lambda$ is the weight for the loss $\mathcal{L}_{supp}$. In this process, the color weight $w_{p}^{t_{0}}$ is not included for the loss. 
Minimizing the changes of initial ray to zero value, we ensure that not just the initial ray but also the warped rays do not diverge, providing a regularization effect.

%
\section{Experiments}
\label{sec:exp}

\subsection{Implementation Details }
\label{subsec:details}

\begin{table}[!t] 
	\begin{center}
		\caption{Quantitative results on synthetic and real-world scenes.}
		\vspace{-2mm}
		\label{tab:comparison}
		\resizebox{\columnwidth}{!}{
		\centering
		\setlength{\tabcolsep}{1pt}
		\renewcommand{\arraystretch}{1.1}
		\scriptsize
		\begin{tabular}{l||c|c|c|c|c|c}
			\toprule 
			
			\multirow{2}{*}{Methods} 			   	& \multicolumn{3}{c|}{Synthetic Scene Dataset}  	   & \multicolumn{3}{c}{Real-World Scene Dataset}  	   \\ \cmidrule{2-6}
			&PSNR($\uparrow$)    &SSIM($\uparrow$)    &LPIPS($\downarrow$) &PSNR($\uparrow$)    &SSIM($\uparrow$)    &LPIPS($\downarrow$)    \\ \midrule \midrule
			Naive NeRF~\cite{mildenhall2020nerf}                            				 & 23.78    & 0.6807   & 0.3362 		& 22.69       & 0.6347      & 0.3687 	\\
			NeRF+MPR~\cite{zamir2021mpr}                           				& 25.11    & 0.7476   & 0.2148 			& 23.38       & 0.6655      & 0.3140      	  	\\
			NeRF+PVD~\cite{son2021pvd}							  			& 24.58 & 0.7190 & 0.2475  & 23.10  & 0.6389 & 0.3425 \\ \midrule
			Deblur-NeRF~\cite{ma2022deblurnerf}                          			& 28.77    & 0.8593   & 0.1400 		 & 25.63       & 0.7645      & 0.1820 		\\
			PDRF-10*~\cite{peng2023pdrf}                         					   & 28.86   & \cellcolor{second!35}0.8795   & 0.1139 		 		  & 25.90       & 0.7734      & 0.1825        					  \\
			BAD-NeRF*~\cite{wang2023bad}										   & 27.32   & 0.8178   & \cellcolor{second!35}0.1127 		 		  & 22.82       & 0.6315      & 0.2887        					\\
			DP-NeRF~\cite{lee2023dp}                        						  & \cellcolor{second!35}29.23    & 0.8674   & 0.1184 		& \cellcolor{second!35}25.91   	 & \cellcolor{second!35}0.7751      & \cellcolor{second!35}0.1602    		   \\ \midrule \midrule
			\textbf{SMURF}                       						   & \cellcolor{best!25}30.98    & \cellcolor{best!25}0.9147   & \cellcolor{best!25}0.0609 		 & \cellcolor{best!25}26.52   		& \cellcolor{best!25}0.7986      & \cellcolor{best!25}0.1013      \\ \bottomrule
		\end{tabular}}
	\end{center}
	\vspace{-6mm}
\end{table}

\paragraph{\textbf{Datasets. }}

In this experiment, we utilize the camera motion blur dataset published by~\cite{ma2022deblurnerf}, which is divided into synthetic and real-world scenes. The synthetic scene dataset comprises five scenes synthesized using Blender~\cite{blender}, with multi-view cameras set up to render motion-blurred images.
The real-world scene dataset consists of ten scenes captured with an actual camera. The blurry images are obtained by physically shaking the camera during the exposure time, and the camera poses are calibrated by COLMAP~\cite{schonberger2016pixelwise,schonberger2016structure}.

\subsection{Novel View Synthesis Results}
\label{sec:results}

We show the quantitative evaluation of our network, SMURF, in comparison to various baselines on the two datasets as shown in~\cref{tab:comparison}. We demonstrate superior quantitative performance across all the metrics. For qualitative results, we compare our results with previous 3D scene deblurring methods in~\cref{fig:novel_view}.

\begin{table}[!h] 
	\begin{center}
		\caption{Ablations on embedding type and regularization strategies. ``Emb.'', ``O.S. Loss'', and ``R.M.'' refer to embedding type, the output suppression loss, and the residual momentum, respectively. ``C.V'' and ``T'' denote chrono-view and time embeddings.}
		\vspace{-2mm}
		\label{tab:regular}
		\resizebox{\columnwidth}{!}{
		\begin{tabular}{c|c|c|c|c|c|c|c|c}
			\toprule 
 \multirow{2}{*}{Emb.} & \multirow{2}{*}{O.S.} & \multirow{2}{*}{R.M.}			   	& \multicolumn{3}{c|}{Synthetic Dataset}  	   & \multicolumn{3}{c}{Real-World Dataset}  \\ \cmidrule{4-9}
			 &  &  & PSNR    &SSIM    &LPIPS &PSNR    &SSIM    &LPIPS     	\\ \midrule \midrule
						C.V & & 				& 26.17   & 0.8186  & 0.0759 			&  25.27      & 0.7576      & 0.1121      	\\
			C.V & & \ding{51}				&  28.49  & 0.8727  & 0.0675  			&  25.60      & 0.7748      & 0.1057      	\\
			C.V  & \ding{51} & 		&  30.72 & 0.9052 &  0.0684          & 25.99  & 0.7846 & 0.1181 \\ \midrule
			 --	& \ding{51} & \ding{51} & 30.54   & 0.8995  & 0.0709 			& 25.74       & 0.7755      & 0.1194      	\\
			 	T.		& \ding{51}  & \ding{51} & 30.94   & \textbf{0.9158}  & 0.0610 			& 26.34       & 0.7881      & 0.1102      	\\ 
			 C.V & \ding{51} & \ding{51}		& \textbf{30.98} & 0.9147 & \textbf{0.0609}           & \textbf{26.52}  & \textbf{0.7986} & \textbf{0.1013} \\ \bottomrule
		\end{tabular}
	}
	\end{center}
	\vspace{-7mm}
\end{table}

\subsection{Ablation Study}
\label{sec:abl}


\paragraph{\textbf{Chrono-View Embedding Function. }}

We conduct experiments by dividing the embedding types into three categories. The time embedding assumes that all images have the same exposure time without incorporating view information, while the chrono-view embedding applies view information, ensuring that all images have distinct exposure time. As shown in~\cref{tab:regular}, the chrono-view embedding shows higher performance in real-world scenes, while its effect in synthetic scenes is minimal. This is attributed to the characteristics of the synthetic dataset created by Deblur-NeRF using Blender, where motion blur images are acquired by linearly interpolating between camera poses. In other words, all the images in the synthetic scenes set share the same exposure time. Our chrono-view embedding function assumes that all images have varying exposure times, leading to its minimal effect on the synthetic dataset. However, the real-world dataset consists of images captured with an actual camera, where the speed of camera motion is not constant during the exposure time. Therefore, applying this function to the real-world scenes yield improved performance due to these variations.

		%

\paragraph{\textbf{Regularzation Strategies. }}

We conduct various experiments for regularization strategies as shown in~\cref{tab:regular}. The output suppression loss yields higher performance as the changes in rays are prevented from diverging since the estimated poses also share the same derivative function \(f\). 
For residual momentum, the sequential camera pose does not deviate significantly from the previous one.~\cref{tab:regular} shows that applying residual momentum shows better performance in terms of LPIPS. 

		%

\section{Conclusion}
\label{sec:con}

We have proposed SMURF, a novel approach that sequentially models camera motion for reconstructing sharp 3D scenes from motion-blurred images. SMURF incorporates a kernel for estimating sequential camera motions, named the CMBK with by two regularization techniques: residual momentum and output suppression loss. We model the 3D scene using TensoRF, which allows for the integration of blurred information and sharp information within adjacent voxels. SMURF significantly outperform previous works quantitatively and qualitatively.
{
    \small
    \bibliographystyle{ieeenat_fullname}
    \bibliography{main}
}


\end{document}